\documentclass[review,times,3p, 11pt]{elsarticle}

\usepackage{enumitem}
\usepackage{lineno}
\usepackage[colorlinks,linkcolor=red,anchorcolor=blue,citecolor=green]{hyperref}
\usepackage{multirow}
\usepackage{rotating}
\usepackage{epstopdf}
\usepackage{graphicx}
\usepackage{subfigure}

\usepackage{times}

\usepackage{makeidx}
\usepackage{multirow}
\usepackage{multicol}
\usepackage[dvipsnames,svgnames,table]{xcolor}
\usepackage{graphicx}
\usepackage{epstopdf}
\usepackage{ulem}

\usepackage{array}
\usepackage{amsmath,amssymb,amsfonts}
\usepackage{algorithmic}
\usepackage{algorithm}
\usepackage{graphicx}
\usepackage{textcomp}

\usepackage{epsfig}
\usepackage{makecell}
\usepackage{longtable,booktabs}
\usepackage{longtable}

\usepackage{lscape}

\journal{Computer Science Review}

\begin{document}
\begin{frontmatter}
	
	
		
	\title{Julia Language in Machine Learning: Algorithms, Applications, and Open Issues}


\author[label1]{Kaifeng Gao}
\author[label1]{Gang Mei\corref{cor1}}
\ead{gang.mei@cugb.edu.cn}
\cortext[cor1]{Corresponding author}
\author[label2]{Francesco Piccialli\corref{cor1}}
\ead{francesco.piccialli@unina.it}
\author[label2]{Salvatore Cuomo\corref{cor1}}
\ead{salvatore.cuomo@unina.it}
\author[label1]{Jingzhi Tu}
\author[label1]{Zenan Huo}
\address[label1]{School of Engineering and Technology, China University of Geosciences (Beijing), 100083, Beijing, China}
\address[label2]{Department of Mathematics and Applications R. Caccioppoli, University of Naples Federico II, Naples, Italy}

\begin{abstract}
Machine learning is driving development across many fields in science and engineering. A simple and efficient programming language could accelerate applications of machine learning in various fields. Currently, the programming languages most commonly used to develop machine learning algorithms include Python, MATLAB, and C/C ++. However, none of these languages well balance both efficiency and simplicity. The Julia language is a fast, easy-to-use, and open-source programming language that was originally designed for high-performance computing, which can well balance the efficiency and simplicity. This paper summarizes the related research work and developments in the applications of the Julia language in machine learning. 
It first surveys the popular machine learning algorithms that are developed in the Julia language. 
Then, it investigates applications of the machine learning algorithms implemented with the Julia language.  
Finally, it discusses the open issues and the potential future directions that arise in the use of the Julia language in machine learning.  

\end{abstract}

\end{frontmatter}

\tableofcontents


\newpage
\section*{List of Abbreviations }
\begin{table}[htbp]
	\begin{tabular}{p{35pt}p{165pt}}
		AD  &  Algorithmic Differentiation \\
		APIs& Application Programming Interfaces \\
		CNN  & Convolutional Neural Network  \\
		DPMM& Dirichlet Process Mixture Model \\
		ELM  &  Extreme Learning Machine  \\
		FFGs&	Forney-style Factor Graphs \\
		GBDT&	Gradient-Boosting Decision Tree  \\
		GMMs& Gaussian Mixture Models  \\
		GPU & Graphics Processing Unit  \\
		ICA& Independent Component Analysis  \\	
		IoT& Internet of Things   \\
		JIT& Just-In-Time \\	
		\textit{k}NN& \textit{k}-Nearest Neighbors \\
		LLVM& Low-Level Virtual Machine \\	
		NLP& Natural Language Processing \\
		ODPS& Open Data Processing Service \\	
		PCA& Principal Component Analysis \\
		RNN &  Recurrent Neural Network \\
		SVD& Singular Value Decomposition \\
		SVM& Support Vector Machine \\	
	\end{tabular}
\end{table}

\newpage

\section{Introduction}
\label{sec1}

\begin{figure}[!ht]
	\centering
	\includegraphics[width=0.9\textwidth]{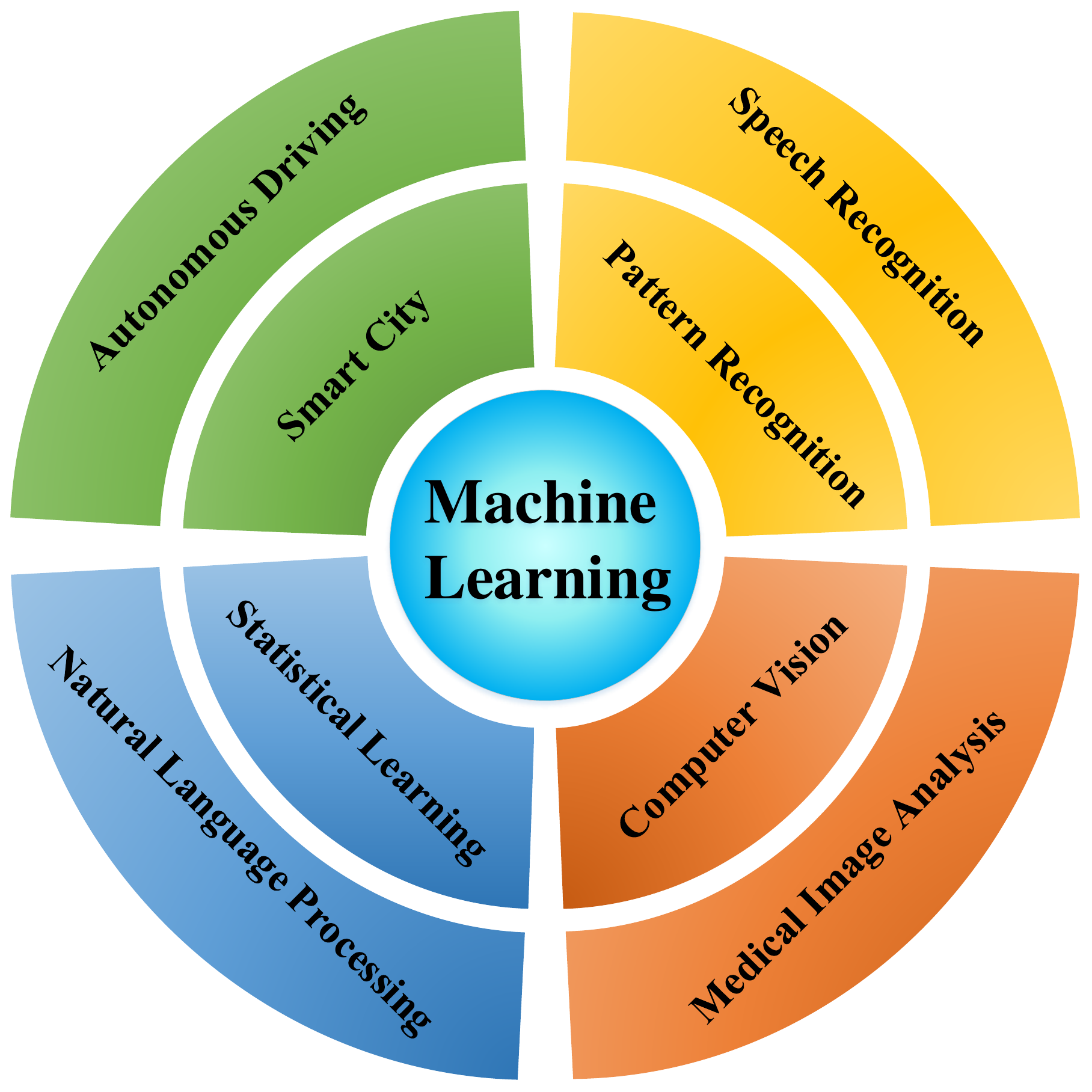}
	\caption{Main applications of machine learning.} \label{fig1}
\end{figure}

Machine learning is currently one of the most rapidly growing technical fields, lying at the intersection of computer science and statistics and at the core of artificial intelligence and data science \cite{1,2,3,4}. Machine learning technology powers many aspects of modern society, from web searches to content filtering on social networks to recommendations on electronic commerce websites. Recent advances in machine learning methods promise powerful new tools for practicing scientists. Modern machine learning methods are closely related to scientific application \cite{5,6}; see Figure \ref{fig1}.

Python, MATLAB, Go, R, and C/C++ are widely used programming languages in machine learning. Python has proven to be a very effective programming language and is used in many scientific computing applications \cite{7}. MATLAB combines the functions of numerical analysis, matrix calculation, and scientific data visualization in an easy-to-use manner. Both Python and MATLAB are "Plug-and-Play" programming languages; the algorithms are prepackaged and mostly do not require learning processes, but they are used to solve large-scale tasks at a slow speed and have very strict requirements for memory and computing power \cite{8}. In addition, MATLAB is commercial.

Go is an open-source programming language that makes it easy to build simple, reliable, and efficient software. Go is syntactically similar to C, but with memory safety, garbage collection, and structural typing. Rather than call out to libraries written in other languages, developers can work with machine learning libraries written directly in Go. However, the current machine learning libraries written in Go are not extensive. R is a language and environment for statistical computing and graphics. R provides a wide variety of statistical and graphical techniques, and is highly extensible. One of the advantages of R is that it can easily produce high-quality drawings. However, R stores data in system memory (RAM), which is a constraint when analyzing big data.

C/C++ is one of the main programming languages in machine learning. It is of high efficiency and strong portability. However, the development and implementation of machine learning algorithms with C/C++ is not easy due to the difficulties in learning and using C/C++. In machine learning, the availability of large data sets is increasing, and the demand for general large-scale parallel analysis tools is also increasing \cite{9}. Therefore, it is necessary to choose a programming language with both simplicity and good performance.

Julia is a simple, fast, and open-source language \cite{10}. The efficiency of Julia is almost comparable to that of static programming languages such as C/C++ and Fortran \cite{11}. Julia is rapidly becoming a highly competitive language in data science and general scientific computing. Julia is as easy to use as R, Python, and MATLAB. 

Julia was originally designed for high-performance scientific computing and data analysis. Julia can call many other mature high-performance basic codes, such as linear algebra and fast Fourier transforms. Similarly, Julia can call C++ language functions directly without packaging or special application programming interfaces (APIs). In addition, Julia has special designs for parallel computing and distributed computing. In high-dimensional computing, Julia has more advantages than C++ \cite{9}. In the field of machine learning, Julia has developed many third-party libraries, including some for machine learning.
 
In this paper, we systematically review and summarize the development of the Julia programming language in the field of machine learning by focusing on the following three aspects:

(1)	Machine learning algorithms developed in the Julia language.
 
(2)	Applications of the machine learning algorithms implemented with the Julia language. 

(3)	Open issues that arise in the use of the Julia language in machine learning. 

The rest of the paper is organized as follows. Section 2 gives a brief introduction to the Julia language. Section 3 summarizes the machine learning algorithms developed in Julia language. Section 4 introduces applications of the machine learning algorithms implemented with Julia language. Section 5 presents open issues occurring in the use of Julia language in machine learning. Finally, Section 6 concludes this survey.
 
\section{A Brief Introduction to the Julia Language}

Julia is a modern, expressive, and high-performance programming language for scientific computing and data processing. Its development started in 2009, and the current stable release as of April 2020 is v1.4.0. Although this low version number indicates that the language is still developing rapidly, it is stable enough to enable the development of research code. Julia's grammar is as readable as that of MATLAB or Python, and it can approach the C/C++ language in performance by compiling in real time. In addition, Julia is a free, open-source language that runs on all popular operation systems.

\begin{figure}[!ht]
	\centering
	\includegraphics[width=0.99\textwidth]{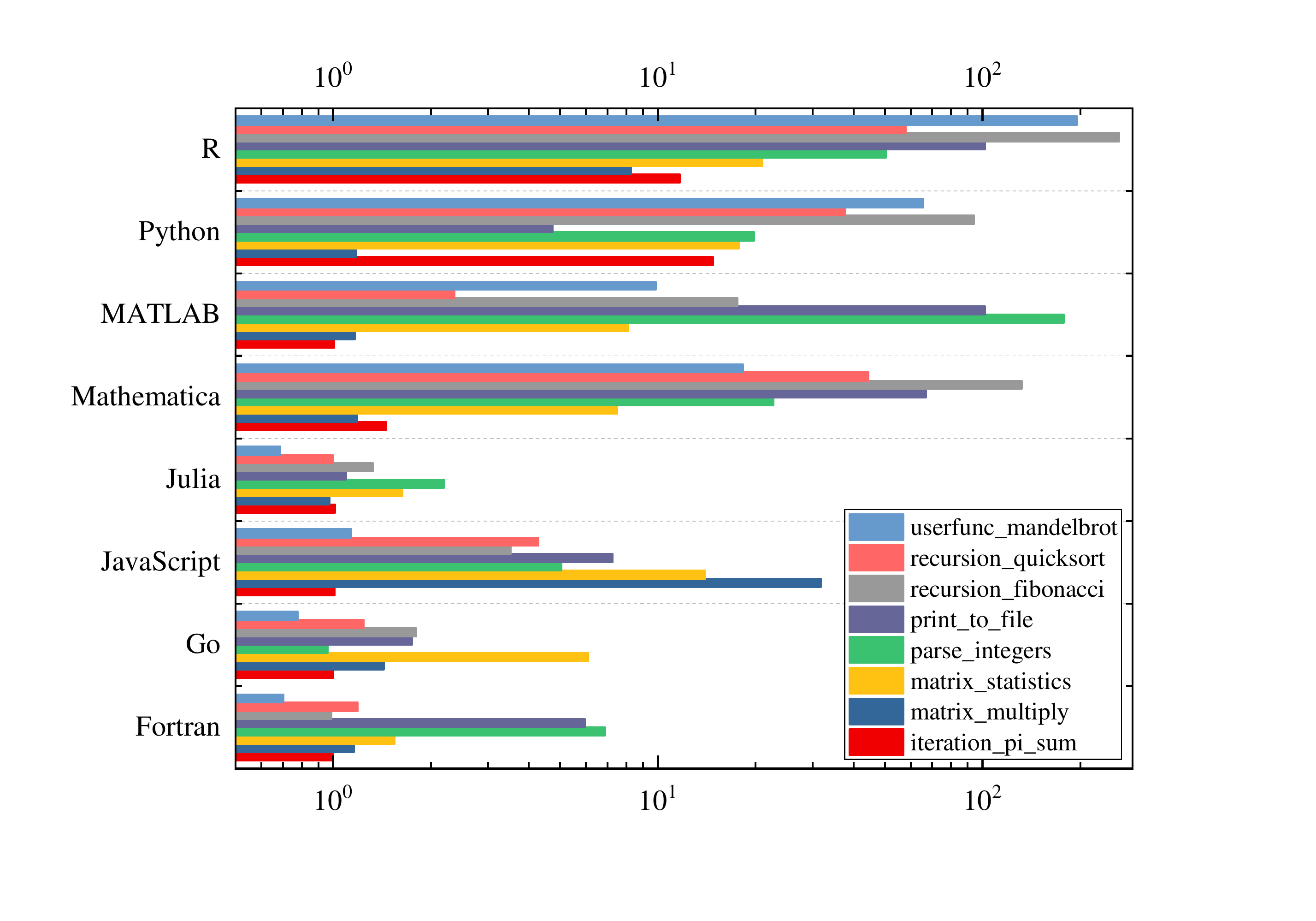}
	\caption{Julia benchmarks (the benchmark data shown above were computed with Julia v1.0.0, Go 1.9, Javascript V8 6.2.414.54, MATLAB R2018a, Anaconda Python 3.6.3, and R 3.5.0. C and Fortran are compiled with gcc 7.3.1, taking the best timing from all optimization levels. C performance = 1.0, smaller is better \cite{15}.)} \label{fig2}
\end{figure}

With the low-level virtual machine (LLVM)-based just-in-time (JIT) compiler, Julia provides powerful computing performance. \cite{12,13}; see Figure \ref{fig2}. Julia also incorporates some important features from the beginning of its design, such as excellent support for parallelism \cite{14} and a practical functional programming orientation, which were not fully implemented in the development of scientific computing languages decades ago. Julia can also be embedded in other programming languages. These advantages make Julia a universal language.

Julia successfully combines the high performance of a static programming language with the flexibility of a dynamic programming language \cite{13}. It provides built-in primitives for parallel computing at every level: instruction level parallelism, multi-threading and distributed computing. The Julia modules allow users to suspend and resume computations with full control of communication without having to manually interface with the operating system's scheduler. Besides, Julia provides a multiprocessing environment based on message passing to allow programs to run on multiple processes in separate memory domains at once \cite{2101}. Moreover, the use of the high-level Julia programming language enables new and dynamic approaches for graphics processing unit (GPU) programming, and Julia GPU code can be highly generic and flexible, without sacrificing performance \cite{14,2103}. However, parallel computing has not yet reached the required level of richness and interactivity \cite{10}. The Julia language could further improve the efficiency of data division and combination, and optimize the parallel algorithms.

\section{Julia in Machine Learning: Algorithms}

\subsection{Overview}

\begin{figure}[!ht]
	\centering
	\includegraphics[width=1\textwidth]{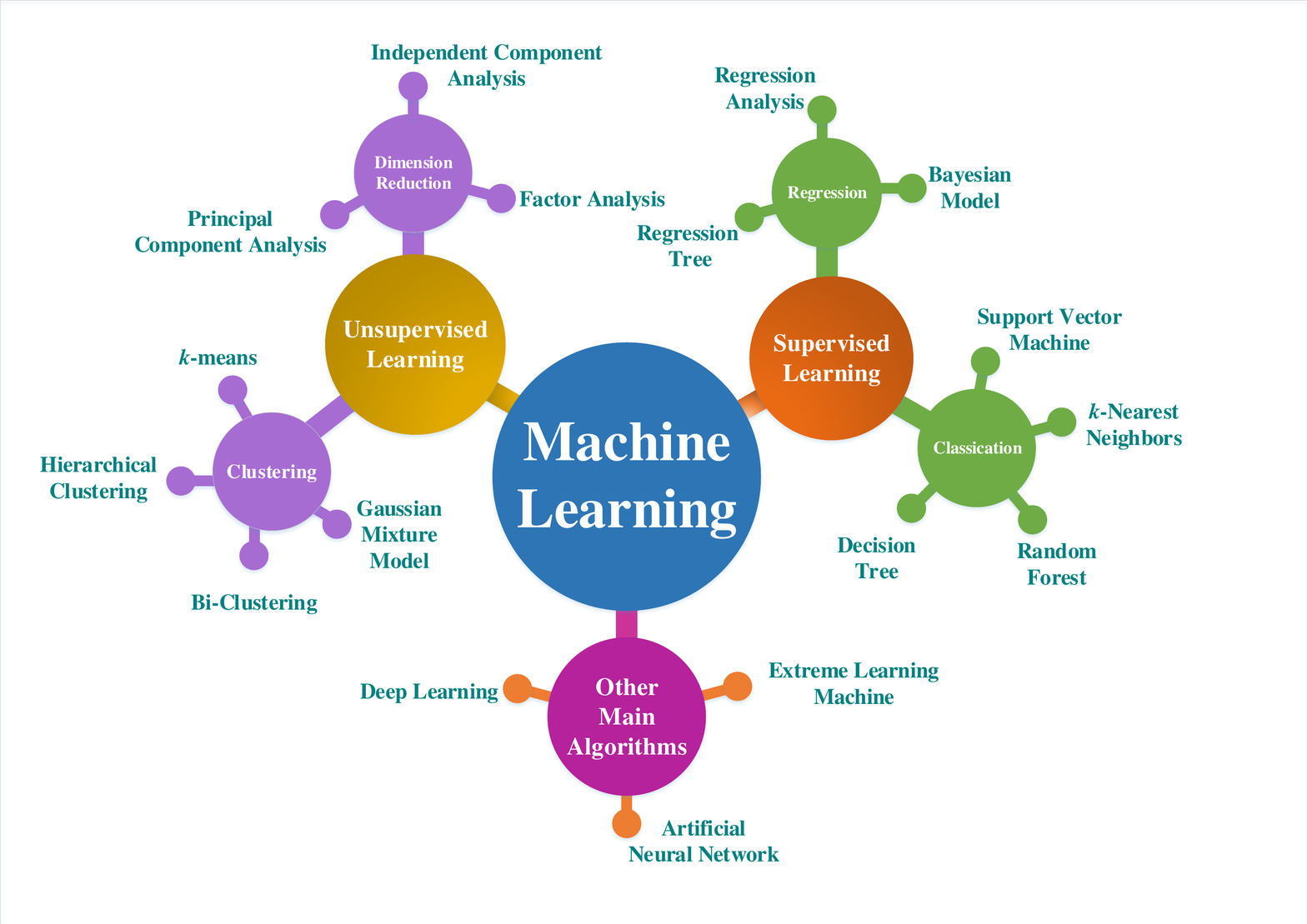}
	\caption{Main machine learning algorithms} \label{fig3}
\end{figure}

This section describes machine learning algorithm packages and toolkits written either in or for Julia. Most applications of machine learning algorithms in Julia can be divided into supervised learning and unsupervised learning algorithms. However, more complex algorithms, such as deep learning, artificial neural networks, and extreme learning machines, include both supervised learning and unsupervised learning, and these require separate classification; see Figure \ref{fig3}.

Supervised learning learns the training samples with class labels and then predicts the classes of data outside the training samples. All the markers in supervised learning are known; therefore, the training samples have low ambiguity. Unsupervised learning learns the training samples without class labels to discover the structural knowledge in the training sample set. All categories in unsupervised learning are unknown; thus, the training samples are highly ambiguous. 

\subsection{Supervised Learning Algorithms Developed in Julia}

\begin{figure}[!ht]
	\centering
	\includegraphics[width=0.9\textwidth]{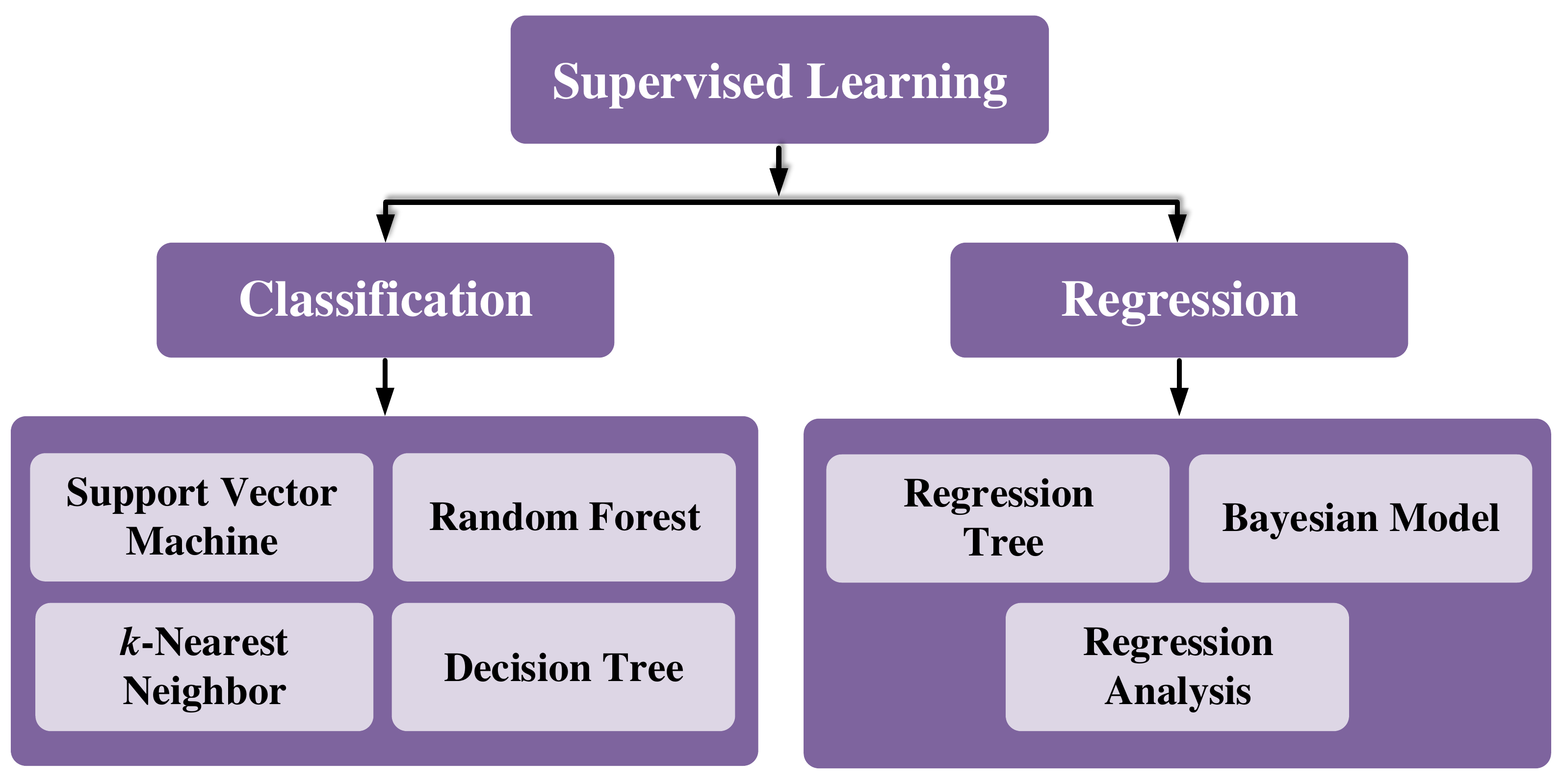}
	\caption{Main supervised learning algorithms developed in Julia} \label{fig4}
\end{figure}

Supervised learning infers a model from labeled training data. Supervised learning algorithms developed in Julia mainly include classification and regression algorithms; see Figure \ref{fig4}.

\subsubsection*{Bayesian Model}

There are two key points in the definition of Bayesian model: independence between features and the Bayesian theorem. One of the most important research areas of Bayesian model is Bayesian linear regression. Bayesian linear regression solves the problem of overfitting in maximum likelihood estimation. Moreover, it makes full use of data samples and is suitable for modeling complex data \cite{1031,1032}. In addition to regression, Bayesian reasoning can also be applied in other fields. Some researchers have conducted research on naive Bayes in image recognition and text classification.

There are some Bayesian model packages and algorithms developed in mature languages. Strickland et al. \cite{16} developed the Python package \textbf{Pyssm}, which was developed for time series analysis using a linear Gaussian state-space model. Mertens et al. \cite{17} developed a user-friendly Python package \textbf{Abrox} for approximate Bayesian computation with a focus on model comparison. There are also Python packages \textbf{BAMSE} \cite{18}, \textbf{BayesPy} \cite{19}, \textbf{PyMC} \cite{20} and so on. Moreover, Vanhatalo et al. \cite{21} developed the MATLAB toolbox \textbf{GPstuff} for Bayesian modeling with Gaussian processes, and Zhang et al. \cite{22} developed the MATLAB toolbox \textbf{BSmac}, which implements a Bayesian spatial model for brain activation and connectivity.

The Julia language is also used to develop packages for the Bayesian model. \textbf{Gen} \cite{23} is a probabilistic programming language proposed by Cusumano and Mansinghka that can be embedded in Julia. This language provides a structure for the optimization of the automatic generation of custom reasoning strategies for static analysis based on an objective probability model. They described \textbf{Gen}'s language design informally and used an example Bayesian statistical model for robust regression to show that \textbf{Gen} is more expressive than \textbf{Stan}, a widely used language for hierarchical Bayesian modeling. Cox et al. \cite{24} explored a specific probabilistic programming paradigm, namely, message passing in Forney-style factor graphs (FFGs), in the context of the automated design of efficient Bayesian signal processing algorithms. Moreover, they developed \textbf{ForneyLab.jl} as a Julia Toolbox for message passing-based inference in FFGs.

Due to the increasing availability of large data sets, the need for a general-purpose massively parallel analysis tool is becoming ever greater. Bayesian nonparametric mixture models, exemplified by the Dirichlet process mixture model (DPMM), provide a principled Bayesian approach to adapt model complexity to the data. Dinari et al. \cite{9} used Julia to implement efficient and easily modifiable distributed inference in DPMMs.

\subsubsection*{\textit{k}-Nearest Neighbors (\textit{k}NN)}

The \textit{k}NN algorithm has been widely used in data mining and machine learning due to its simple implementation and distinguished performance. A training data set with a known label category is used, and for a new data set, the \textit{k} instances closest to the new data are found in the feature space of the training data set. If most of the instances belong to a category, the new data set belongs to this category.

At present, there are many packages developed for the \textit{k}NN algorithm in the Python language. Among these, \textbf{scikit-learn} and \textbf{Pypl} are the most commonly used packages. It should be noted that \textbf{scikit-learn} and \textbf{Pypl} are not specially developed for the \textit{k}NN algorithm; they contain many other machine learning algorithms. In addition, Bergstra et al. \cite{25} developed \textbf{Hyperopt} to define a search space that encompasses many standard components and common patterns of composing them.

Julia is also used to develop packages for the \textit{k}NN algorithm. \textbf{NearestNeighbors.jl} \cite{26} is a package written in Julia to perform high-performance nearest neighbor searches in arbitrarily high dimensions. This package can realize \textit{k}NN searches and range searches.

\subsubsection*{Decision Tree, Regression Tree, and Random Forest}

Mathematically, a decision tree is a graph that evaluates a limited number of probabilities to determine a reliable classification for each data point. A regression tree is the opposite of a decision tree and is suitable for solving regression problems. It does not predict labels but predicts a continuous change value. Random forests are a set of decision trees or regression trees that work together \cite{27}. The set of decision trees (or continuous y regression trees) is constructed by performing bootstrapping on the data sets and averaging or acquiring pattern prediction (called "bagging") from the trees. Subsampling of features is used to reduce generalization errors \cite{28}. An ancillary result of the bootstrapping procedure is that the data not sampled in each bootstrap (called "out-of-bag" data) can be used to estimate the generalization error as an alternative to cross-validation \cite{29}.

Many packages have been developed for decision trees, regression trees, and random forests. For example, the above three algorithms are implemented in \textbf{Spark2 ML} and \textbf{scikit-learn} using Python. In addition, Upadhyay et al. \cite{30} proposed land-use and land-cover classification technology based on decision trees and \textit{k}-nearest neighbors, and the proposed techniques are implemented using the \textbf{scikit-learn} data mining package for python. Keck \cite{31} proposed a speed-optimized and cache-friendly implementation for multivariate classification called \textbf{FastBDT}, which provides interfaces to C/C++, Python, and TMVA. Yang et al. \cite{32} used the open data processing service (ODPS) and Python to implement the gradient-boosting decision tree (GBDT) model.

\textbf{DecisionTree.jl} \cite{33}, written in the Julia language, is a powerful package that can realize decision tree, regression tree, and random forest algorithms very well. The package has two functions, and the ingenious use of these functions can help us realize these three algorithms.

\subsubsection*{Support Vector Machine (SVM)}

In SVM, the objective is to find a hyperplane in high-dimensional space, which represents the maximum margin between any two instances of two types of training data points (support vectors) or maximizes the correlation function when it cannot be separated. The so-called kernel similarity function is used to design the non-linear SVM \cite{34}. 

Currently, there are textbook style implementations of two popular linear SVM algorithms: Pegasos \cite{35}, Dual Coordinate Descent. \textbf{LIBSVM} developed by the Information Engineering Institute of Taiwan University is the most widely used SVM tool \cite{36}. \textbf{LIBSVM} includes standard SVM algorithm, probability output, support vector regression, multi-classification SVM and other functions. Its source code is originally written by C. It provides Java, Python, R, MATLAB, and other language invocation interfaces. 

\textbf{SVM.jl} \cite{37}, \textbf{MLJ.jl} \cite{38}, and \textbf{LIBSVM.jl} \cite{39} are native Julia implementations of SVM algorithm. However, \textbf{LIBSVM.jl} is more comprehensive than \textbf{SVM.jl}. \textbf{LIBSVM.jl} supports all libsvm models: classification c-svc, nu-svc, regression: epsilon-svr, nu-svr and distribution estimation: a class of support vector machines and \textbf{ScikitLearn.jl} \cite{40} API. In addition, the model object is represented by a support vector machine of Julia type. The SVM can easily access the model features and can be saved as a JLD file.

\subsubsection*{Regression Analysis}

Regression analysis is an important supervised learning algorithm in machine learning. It is a predictive modeling technique, which constructs the optimal solution to estimate unknown data through the sample and weight calculation. Regression analysis is widely used in the fields of the stock market and medical data analysis.

Python has been widely used to develop a variety of third-party packages for regression analysis, including \textbf{scikit-learn} and \textbf{orange}. The \textbf{scikit-learn} package is a powerful Python module, which supports mainstream machine learning algorithms such as regression, clustering, classification and neural network \cite{41,42,43}. The \textbf{orange} package is a component-based data mining software, which can be used as a module of Python programming language, especially suitable for classification, clustering, regression and other work \cite{44,45}. MATLAB also supports the regression algorithm. By invoking commands such as \textit{regress} and \textit{stepwise} in the statistical toolbox of MATLAB, regression operation can be performed conveniently on the computer.

The Julia language is also used to develop a package, \textbf{Regression.jl} \cite{46}, to perform the regression analysis. The \textbf{Regression.jl} package seeks to minimize empirical risk based on \textbf{EmpiricalRisk.jl} \cite{47} and provides a set of algorithms for performing regression analysis. It supports multiple linear regression, non-linear regression, and other regression algorithms. In addition, the \textbf{Regression.jl} package also provides a variety of solvers such as analytical solution (for linear and ridge regression) and gradient descent.

\subsection{Unsupervised Learning Algorithms Developed in Julia}

\begin{figure}[!ht]
	\centering
	\includegraphics[width=0.99\textwidth]{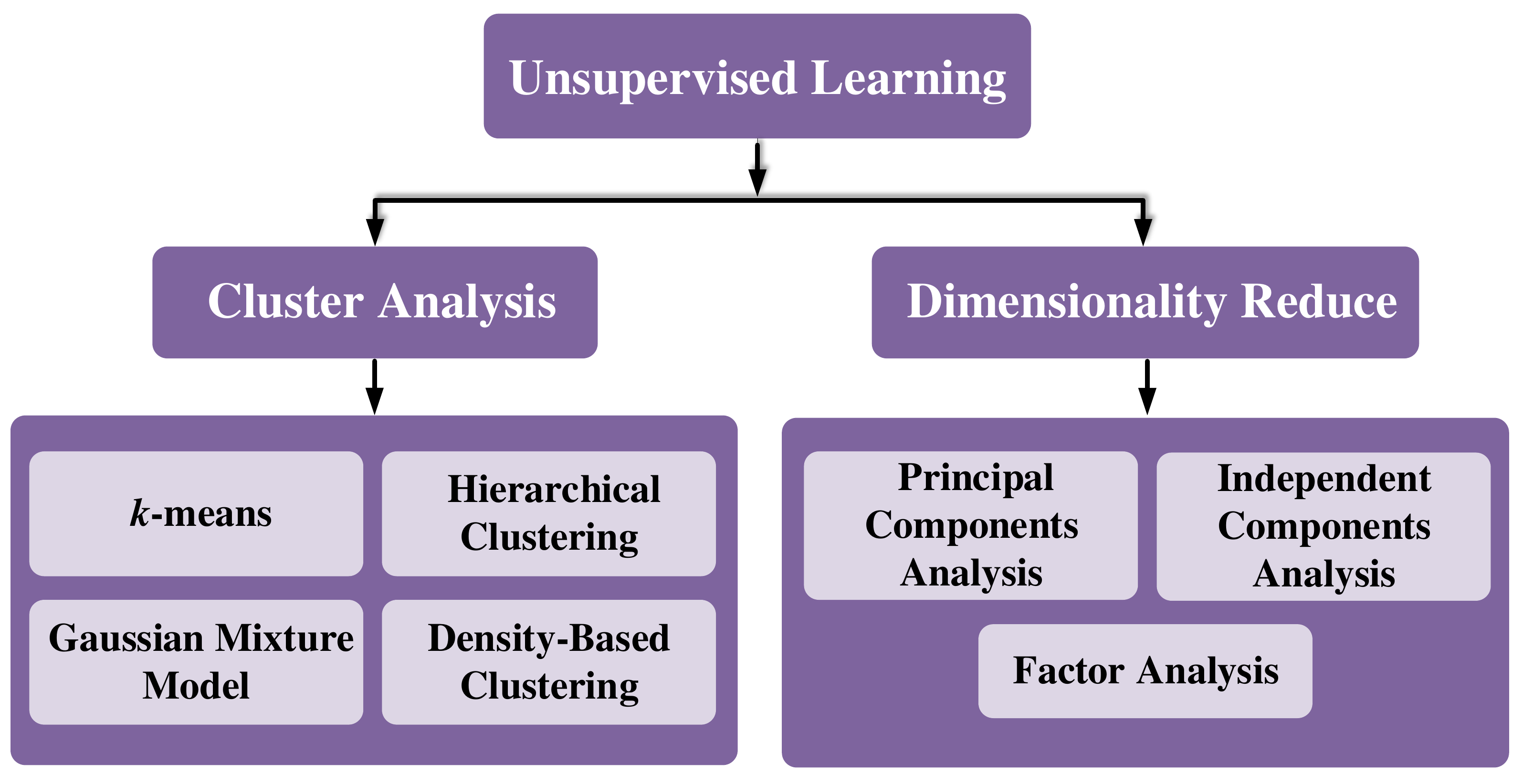}
	\caption{Main unsupervised learning algorithms developed in Julia} \label{fig5}
\end{figure}

Unsupervised learning is a type of self-organized learning that can help find previously unknown patterns in a dataset without the need for pre-existing labels. Two of the main methods used in unsupervised learning are dimensionality reduction and cluster analysis; see Figure \ref{fig5}. 

\subsubsection*{Gaussian Mixture Models (GMMs)}

GMMs are probabilistic models for representing normally distributed subpopulations within an overall population. GMMs use Gaussian distribution as the basic parameter model, accurately characterizes the data distribution by combining multiple Gaussian distributions, and use the expectation-maximization algorithm for training. Compared with the Gaussian models, the GMMs provide greater flexibility and precision in modeling the underlying statistics of sample data \cite{49}. Generally, the GMMs are used to solve problems such as image segmentation and dynamic target detection \cite{48}.

Currently, there are many libraries that can implement Gaussian mixture models; these include packages developed with Python, such as \textbf{PyBGMM} and \textbf{numpy-ml}, and packages developed with C++, such as \textbf{Armadillo}. There are also some GMM packages for specialized fields. Bruneau et al. \cite{50} proposed a new Python package for nucleotide sequence clustering, which implements a Gaussian mixture model for DNA clustering. Holoien et al. \cite{51} developed a new open-source tool, \textbf{EmpiriciSN}, written in Python, for performing extreme deconvolution Gaussian mixture modeling.

To the best of the authors' knowledge, there is no mature Julia package for the GMMs. \textbf{GmmFlow.jl} \cite{2051} is a Julia library that can implement some simple functions of the GMMs, including model generation and cluster mapping. But the algorithm optimization could be improved. \textbf{GaussianMixtures.jl} \cite{2052} is another Julia package for the GMMs. This package has implemented both diagonal covariance and full covariance GMMs, and full covariance variational Bayes GMMs. However, the package is slightly strict with data types. \textbf{ScikitLearn.jl} implements the popular \textbf{scikit-learn} interface and algorithms in Julia, and it can access approximately 150 Julia and Python models, including the Gaussian mixture model. Moreover, Srajer et al. \cite{52} used algorithmic differentiation (AD) tools in a GMM fitting algorithm.

\subsubsection*{\textit{k}-means}

The \textit{k}-means clustering algorithm is an iterative clustering algorithm. It first randomly selects \textit{k} objects as the initial clustering center, then calculates the distance between each object and each seed clustering center, and assigns each object to the nearest clustering center. Cluster centers and the objects assigned to them represent a cluster. As an unsupervised clustering algorithm, \textit{k}-means is widely used because of its simplicity and effectiveness.

The \textit{k}-means algorithm is a classic clustering method, and many programming languages have developed packages related to it. The third-party package \textbf{scikit-learn} in Python implements the \textit{k}-means algorithm \cite{43,53}. The \textit{Kmeans} function in MATLAB can also implement a \textit{k}-means algorithm \cite{54}. In addition, many researchers have implemented the \textit{k}-means algorithm in the C/C++ programming language.

Julia has also been used to develop a specific package, \textbf{Clustering.jl} \cite{55}, for clustering. \textbf{Clustering.jl} provides several functions for data clustering and clustering quality evaluation. Because \textbf{Clustering.jl} has comprehensive and powerful functions, this package is a good choice for \textit{k}-means.

\subsubsection*{Hierarchical Clustering}

Hierarchical clustering is a kind of clustering algorithm that performs clustering by calculating the similarity between data points of different categories \cite{56,57,58}. The strategy of cohesive hierarchical clustering is to first treat each object as a cluster and then merge these clusters into larger and larger clusters until all objects are in one cluster or some termination condition is satisfied.

The commonly used Python packages for hierarchical clustering are \textbf{scikit-learn} and \textbf{scipy}. Hierarchical clustering within the \textbf{scikit-learn} package is implemented in the \textit{sklearn.cluster} method, which includes three important parameters: the number of clusters, the connection method, and connection measurement options \cite{43}. \textbf{scipy} implements hierarchical clustering with the \textit{scipy.cluster} method \cite{59}. In addition, programming languages such as MATLAB and C/C++ can also perform hierarchical clustering \cite{60}.

The package \textbf{QuickShiftClustering.jl} \cite{61}, written using Julia, can realize hierarchical clustering algorithms. This package is quite easy to use. It provides three functions: clustering matrix data, clustering labels, and creating hierarchical links to achieve hierarchical clustering \cite{62}.

\subsubsection*{Bi-Clustering}

Bi-Clustering algorithm is based on traditional clustering. Its basic idea is to cluster rows and columns of matrices through traditional clustering, and then merge the clustering results. Bi-Clustering algorithm solves the bottleneck problem of traditional clustering in high-dimensional data. Data sets in reality are mostly high-dimensional and inherently sparse. Traditional clustering algorithms often fail to detect meaningful clustering in high-dimensional data sets. However, Bi-Clustering can detect clusters of any shape and position in space, and it is an effective method to solve the problem of subspace clustering in high-dimensional data sets \cite{2061,2062}. To search for local information better in the data matrix, researchers put forward the concept of bi-clustering. 

The package \textbf{scikit-learn} can implement bi-clustering, and the implementation module is \textit{sklearn.cluster.bicluster}. At present, bi-clustering is mainly applied to highthroughput detection technologies such as gene chips and DNA microarrays. 

The Julia language is also used to develop packages that implement bi-clustering. For example, \textbf{Kpax3} \cite{kpax3} is a Bayesian method for multi-cluster multi-sequence alignment. Bezanson et al. \cite{10} used a Bayesian dual clustering model, which extended and improved the model originally introduced by Pessia et al. \cite{kpax3}. They wrote the \textbf{kpax3.jl} library package in Julia and the output contains multiple text files containing a cluster of rows and columns of the input dataset. 

\subsubsection*{Principal Component Analysis (PCA)}

PCA is a method of statistical analysis and a simplified data set. It uses an orthogonal transformation to linearly transform observations of a series of possibly related variables and then project them into a series of linearly uncorrelated variables. These uncorrelated variables are called principal components. PCA is often used to reduce the dimensionality of a data set while maintaining the features that have the largest variance contribution in the data set.

Python is the most frequently used language for developing PCA algorithms. The \textbf{scikit-learn} package provides a class, \textit{sklearn.decomposition.PCA} \cite{63}, to implement PCA algorithms in the \textit{sklearn.decomposition} module. Generally, the PCA class does not need to adjust parameters very much but needs to specify the target dimension or the variance of the principal components after dimensionality reduction. In addition, many researchers have developed related application packages using the C++ programming language. These include the \textbf{ALGLIB} \cite{64} package and the class \textbf{cv :: PCA} \cite{65} in OpenCV.

To the best of the authors' knowledge, there is no mature Julia package specifically for PCA. However, \textbf{MultivariateStats.jl} \cite{66} is a Julia package for multivariate statistics and data analysis. This package defines a PCA type to represent a PCA model and provides a set of methods to access properties.

\subsubsection*{Independent Component Analysis (ICA)}

ICA is a new signal processing technology developed in recent years. The ICA method is based on mutual statistical independence between sources. In the practical applications of signal processing, especially in communication and biomedicine, it is important to eliminate noise data. Traditional signal processing techniques for noise cancellation include band-pass filtering, fast Fourier transform, autocorrelation, autoregressive modeling, adaptive filtering, Kalman filtering and singular value decomposition. The traditional filtering technology is based on the assumption that noise is the only additive, which is not suitable for multi-sensor observation of mixed signals \cite{1071}. However, the ICA algorithm has strong robustness to additive noise, and it is one of the most promising methods to solve the problem of blind noise suppression \cite{1072,1073}. Moreover, in contrast to traditional signal separation methods based on feature analysis, such as singular value decomposition (SVD) and PCA, ICA is an analysis method based on higher-order statistical characteristics. In many applications, the analysis of higher-order statistical characteristics is more practical.

Python is the most frequently used language in developing ICA algorithms. The \textbf{scikit-learn} package has developed a class, \textit{FastICA} \cite{67}, to implement ICA algorithms in the \textit{sklearn.decomposition} module. In addition, Brian Moore \cite{68} developed a \textbf{PCA and ICA Package} using the MATLAB programming language. The PCA and ICA algorithms are implemented as functions in this package, and it includes multiple examples to demonstrate their usage.

To the best of the authors' knowledge, ICA does not have a mature software package developed in the Julia language. However, \textbf{MultivariateStats.jl} \cite{66}, like a Julia package for multivariate statistics and data analysis, defines an ICA type representing the ICA model and provides a set of methods to access the attributes.

\subsection{Other Main Algorithms}

In addition to supervised learning algorithms and unsupervised learning algorithms, machine learning algorithms include a class of algorithms that are more complex and cannot be categorized into a specific category. For example, artificial neural networks can implement supervised learning, unsupervised learning, reinforcement learning, and self-learning. Deep learning algorithms are based on artificial neural network algorithms and can perform supervised learning, unsupervised learning, and semisupervised learning. Extreme learning machines were proposed for supervised learning algorithms but were extended to unsupervised learning in subsequent developments.

\subsubsection*{Deep Learning}

Deep learning allows computational models that are composed of multiple processing layers to learn representations of data with multiple levels of abstraction \cite{5}. Several deep learning frameworks, such as the depth neural network, the convolutional neural network, the depth confidence network and the recursive neural network, have been applied to computer vision, speech recognition, natural language processing, image recognition, and bioinformatics and have achieved excellent results.

It has been several years since the birth of deep learning algorithms. Many researchers have improved and developed deep learning algorithms. Python is the most frequently used language in developing deep learning algorithms. For example, \textbf{PyTorch} \cite{69,70} and \textbf{ALiPy} \cite{71} are Python packages with many deep learning algorithms. Moreover, Tang et al. developed \textbf{GCNv2} \cite{72} using C++ and Python, Huang et al. wrote \textbf{Mask Scoring R-CNN} \cite{73} using Python, Hanson and Frazier-Logue compared the Dropout \cite{74} algorithm with the SDR \cite{75} algorithm, and Luo et al. \cite{76} proposed and used Python to write AdaBound (a new adaptive optimization algorithm).

Julia has also been used to develop various deep learning algorithms. For example, AD allows the exact computation of derivatives given only an implementation of an objective function, and Srajer et al. \cite{52} wrote an AD tool and used it in a hand-tracking algorithm.

\textbf{Augmentor} is a software package available in both Python and Julia that provides a high-level API for the expansion of image data using a stochastic, pipeline-based approach that effectively allows images to be sampled from a distribution of augmented images at runtime \cite{77}. To demonstrate the API and to highlight the effectiveness of augmentation on a well-known dataset, a short experiment was performed. In the experiment, the package is used on a convolutional neural network (CNN) \cite{78}.

\textbf{MXNet.jl} \cite{79}, \textbf{Knet.jl} \cite{80}, \textbf{Flux.jl} \cite{81}, and \textbf{TensorFlow.jl} \cite{82} are deep learning frameworks with both efficiency and flexibility. At its core, \textbf{MXNet.jl} contains a dynamic dependency scheduler that automatically parallelizes both symbolic and imperative operations on the fly. \textbf{MXNet.jl} is portable and lightweight, scaling effectively to multiple GPUs and multiple machines. 

\subsubsection*{Artificial Neural Networks}

A neural network is a feedforward network consisting of nodes ("neurons"), each side of which has weights. These allow the network to form a mapping between the input and output \cite{83}. Each neuron that receives input from a previous neuron consists of the following components: the activation, a threshold, the time at which the newly activated activation function is calculated and the output function of the activation output.

At present, the framework of a neural network model is usually developed in C++ or Python. \textbf{DLL} is a machine learning framework written in C++ \cite{84}. It supports a variety of neural network layers and standard backpropagation algorithms. It can train artificial neural networks and CNNs and support basic learning options such as momentum and weight attenuation. \textbf{scikit-learn}, a machine learning library based on Python, also supports neural network models \cite{43}.

Employing the Julia language, \textbf{Diffiqflux.jl} \cite{86} is a package that integrates neural networks and differential equations. Rackauckas et al. \cite{86} described differential equations from the perspective of data science and discuss the complementarity between machine learning models and differential equations. These authors demonstrated the ability to combine \textbf{DifferentialEquations.jl} \cite{87} defined differential equations into \textbf{Flux}-defined neural networks. \textbf{Backpropneuralnet.jl} \cite{52} is an easy-to-use neural network package.

\subsubsection*{Extreme Learning Machine (ELM)}

ELM \cite{88} is a variant of Single Hidden Layer Feedforward Networks (SLFNs). Because its weight is not adjusted iteratively, it deviates greatly. This significantly improves the efficiency when training the neural networks. 

The basic algorithm of ELM and Multi-Layer \cite{89}/Hierarchical \cite{90} ELM have been implemented in \textbf{HP-ELM}. Meanwhile, C/C++, MATLAB, Python and JAVA versions are provided. \textbf{HP-ELM} includes GPU acceleration and memory optimization, which is suitable for large data processing. \textbf{HP-ELM} supports LOO (Leave One Out) and \textit{k}-fold cross-validation to dynamically select the number of hidden layer nodes. The available feature maps include linear function, Sigmoid function, hyperbolic sinusoidal function, and three radial basis functions. 

According to ELM, parameters of hidden nodes or neurons are not only independent of training data but also independent of each other. Standard feedforward neural networks with hidden nodes have universal approximation and separation capabilities. These hidden nodes and their related maps are terminologically ELM random nodes, ELM random neurons or ELM random features. Unlike traditional learning methods, which need to see training data before generating hidden nodes or neuron parameters, ELM could generate hidden nodes or neuron parameters randomly before seeing training data. \textbf{Elm.jl} \cite{91} is an easy-to-use extreme learning machine package. 

\subsection{List of Commonly Used Julia Packages}

We summarize the commonly used Julia language packages and the machine learning algorithms that these packages primarily support; see the investigation in Table \ref{tab1}.

\section{Julia in Machine Learning: Applications}

\subsection{Overview}

Machine learning is one of the fastest-growing technical fields nowadays. It is a cross-cutting field of statistics and computer science \cite{1,2,3}. Machine learning specializes in how computers simulate or implement human learning behaviors. By acquiring new knowledge and skills, the existing knowledge structure is reorganized to improve its performance.

\begin{figure}[!ht]
	\centering
	\includegraphics[width=0.9\textwidth]{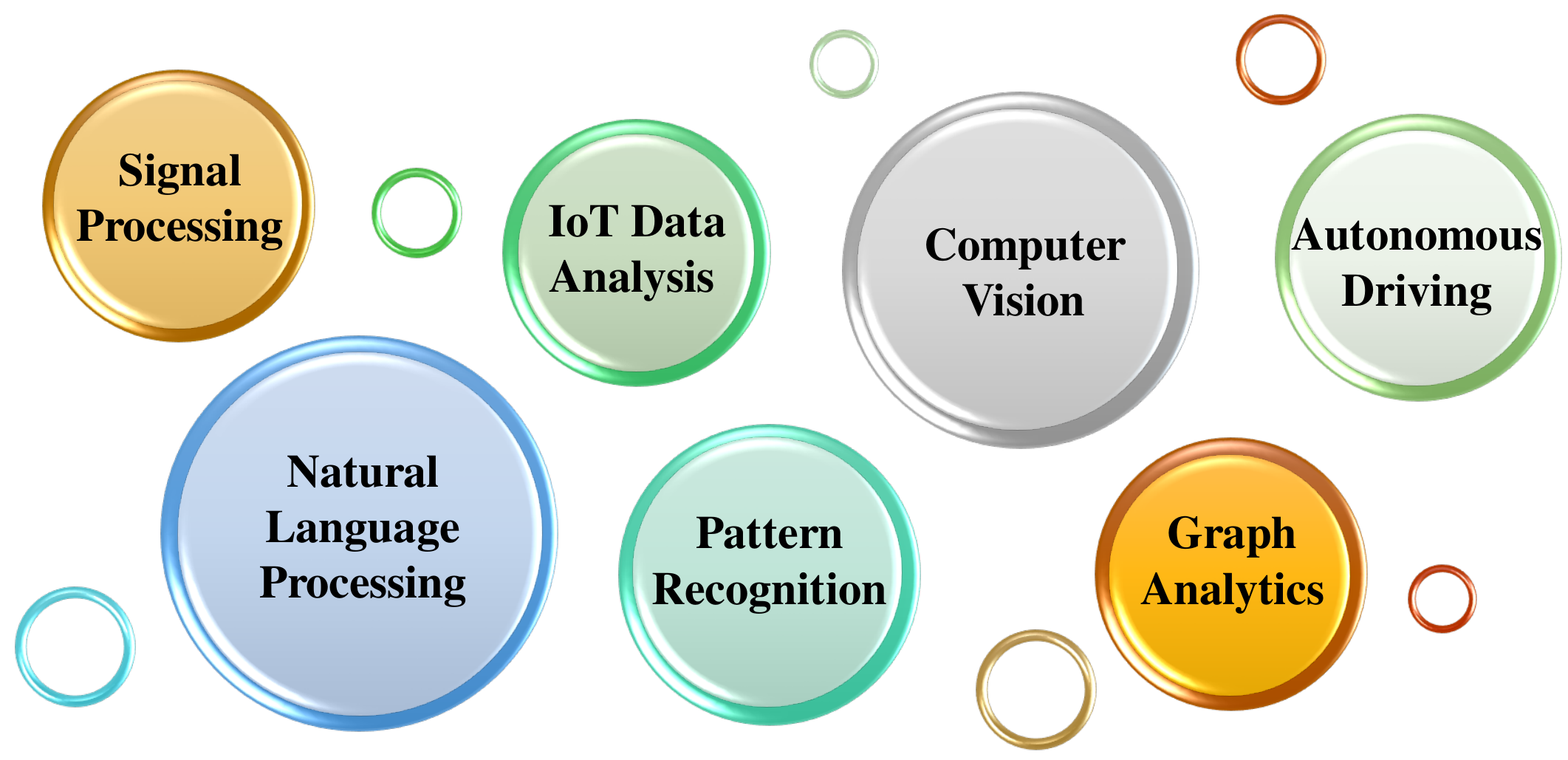}
	\caption{Major applications of machine learning using Julia language} \label{fig6}
\end{figure}

Julia, as a programming language with the rise of machine learning, has corresponding algorithmic library packages in most machine learning applications. In the following, we summarize the applications of Julia in machine learning. As shown in Figure \ref{fig6}, the current applications of Julia programming language in machine learning mainly focus on the Internet of Things (IoT), computer vision, autonomous driving, pattern recognition, etc. 

\subsection{Analysis of IoT Data}

The IoT, also called the Internet of Everything or the Industrial Internet, is a new technology paradigm envisioned as a global network of machines and devices capable of interacting with each other \cite{92}. The application of the IoT in industry, agriculture, the environment, transportation, logistics, security, and other infrastructure fields effectively promotes the intelligent development of these areas and more rationally uses and allocates limited resources, thus improving the efficiency of these fields \cite{93,94,95}. Machine learning has brought enormous development opportunities for the IoT and has a significant impact on existing industries \cite{96,97}. 

Invenia Technical Computing used the Julia language to expand its energy intelligence system \cite{98}. They optimized the entire North American grid and used the energy intelligent system (EIS) and various signals to directly improve the day-ahead planning process. They used the latest research in machine learning, complex systems, risk analysis, and energy systems. In addition, Julia provided Invenia Technical Computing with versatility in terms of programming style, parallelism, and language interoperability \cite{98}.

Fugro Roames engineers \cite{99} used the Julia language to implement machine learning algorithms to identify network faults and potential faults, achieving a 100-fold increase in speed. Protecting the grid means ensuring that all power lines, poles, and wires are in good repair, which used to be a laborious manual task that required thousands of hours to travel along the power line. Fugro Roames engineers have developed a more effective way to identify threats to wires, poles, and conductors. Using a combination of LiDAR and high-resolution aerial photography, they created a detailed three-dimensional map of the physical conditions of the grid and possible intrusions. Then, they used machine learning algorithms to identify points on the network that have failed or are at risk of failure \cite{99}.

\subsection{Computer Vision}

Computer vision is a simulation of biological vision using computers and related equipment. Its main task is to obtain the three-dimensional information of the corresponding scene by processing collected pictures or videos. Computer vision includes image processing and pattern recognition. In addition, it also includes geometric modeling and recognition processes. The realization of image understanding is the ultimate goal of computer vision. Machine learning is developing, and computer vision research has gradually shifted from traditional models to deep learning models represented by CNNs and deep Boltzmann machines.

Computer vision is currently widely applied in the fields of biological and medical image analysis \cite{100}, urban streetscapes \cite{33,101}, rock type identification \cite{102}, automated pavement distress detection and classification \cite{103}, structural damage detection in buildings \cite{104}, and other fields. The development language used in current research is usually Python or another mature language. In contrast, when dealing with large-scale data sets, the Julia language has inherent advantages in high-performance processing. Therefore, many scholars and engineers use Julia to develop packages for the applications of computer vision. The \textbf{Metalhead.jl} \cite{105} package provides computer vision models that run on top of the \textbf{Flux} machine learning library. The package \textbf{ImageProjectiveGeometry.jl} \cite{106} is intended as a starting point for the development of a library of projective geometry functions for computer vision in Julia. Currently, the package consists of a number of components that could ultimately be separated into individual packages or added to other existing packages.

\subsection{Natural Language Processing (NLP)}

NLP employs computational techniques for the purpose of learning, understanding, and producing human language content \cite{107}. It is an important research direction in the field of computer science and artificial intelligence. Modern NLP algorithms are based on machine learning algorithms, especially statistical machine learning algorithms. Many different machine learning algorithms have been applied to NLP tasks, the most representative of which are deep learning algorithms exemplified by CNN \cite{108,109,110,111}.

Currently, one of the main research tasks of NLP is to investigate the characteristics of human language and establish the cognitive mechanism of understanding and generating language. In addition, new practical applications for processing human language through computer intelligence have been developed. Many researchers and engineers have developed practical application tools or software packages using the Julia language. For example, \textbf{LightNLP.jl} \cite{112} is a lightweight NLP toolkit for the Julia language. However, to the best of the authors' knowledge, there are currently no stable library packages developed in the Julia language specifically for NLP.

\subsection{Autonomous Driving}

Machine learning is widely used in autonomous driving, and it mainly focuses on the environmental perception and behavioral decision-making of autonomous vehicles. The application of machine learning in environmental perception belongs to the category of supervised learning. When performing object recognition on images obtained from the surrounding environment of a vehicle, a large number of images with solid objects are required as training data, and then deep learning methods can identify objects from the new images \cite{37,38,91,113}. The application of machine learning in behavioral decision-making generally involves reinforcement learning. Autonomous vehicles need to interact with the environment, and reinforcement learning learns the mapping relationship between the environment and behavior that interacts with the environment from a large amount of sample data. Thus, whenever an autonomous vehicle perceives the environment, it can act intelligently \cite{114,115}.

To the best of the authors' knowledge, there are no software packages or solutions specially developed in Julia for autonomous driving. However, the machine learning algorithms used in autonomous driving are currently implemented by researchers in the Julia language. The amount of data obtained by autonomous vehicles is huge, and the processing is complex, but autonomous vehicles have strict requirements for data processing time. High-level languages such as Python and MATLAB are not as efficient in computing as the Julia language, which was specifically developed for high-performance computing. Therefore, we believe that Julia has strong competitiveness as a programming language for autonomous vehicle platforms.

\subsection{Graph Analytics}

Graph analytics is a rapidly developing research field. It combines graph-theoretic, statistics and database technology to model, store, retrieve and analyze graph-structured data. Samsi \cite{116} used subgraph isomorphism to solve the previous scalability difficulties in machine learning, high-performance computing, and visual analysis. The serial implementations of C++, Python, and Pandas and MATLAB are implemented, and their single-thread performance is measured. 

\textbf{LightGraphs.jl} is currently the most comprehensive library developed in Julia for graph analysis \cite{117}. \textbf{LightGraphs.jl} provides a set of simple, concrete graphical implementations (including undirected and directed graphs) and APIs for developing more complex graphical implementations under the AbstractGraph type. 

\subsection{Signal Processing}

The signal processing in communications is the cornerstones of electrical engineering research and other related fields \cite{118,119}. Python has natural advantages in analyzing complex signal data due to its numerous packages. In addition, the actually collected signals need to be processed before they can be used for analysis. MATLAB provides many signal processing toolboxes, such as spectrum analysis toolbox, waveform viewer, filter design toolbox. Therefore, MATLAB is also a practical tool for signal data processing. 

Current and emerging means of communication increasingly rely on the ability to extract patterns from large data sets to support reasoning and decision-making using machine learning algorithms. This calls the use of the Julia language. For example, Srivastava Prakalp et al. \cite{120} designed an end-to-end programmable hybrid signal accelerator, \textbf{PROMISE}, for machine learning algorithms. \textbf{PROMISE} can use machine learning algorithms described by Julia and generate \textbf{PROMISE} code. \textbf{PROMISE} can combine multiple signals and accelerate machine learning algorithms. 

\subsection{Pattern Recognition}

Pattern recognition is the automatic processing and interpretation of patterns by means of a computer using mathematical technology \cite{121}. With the development of computer technology, it is possible for humans to study the complex process of information-processing, an important form of which is the recognition of the environment and objects by living organisms. The main research directions of pattern recognition are image processing and computer vision, speech information processing, medical diagnosis and biometric authentication technology \cite{122}. 

Pattern recognition is generally categorized according to the type of learning procedure used to generate the output value. Medical diagnosis is a typical field of pattern recognition applications. Rajsavi et al. \cite{123} used the Julia libraries packages such as \textbf{GLM.jl} \cite{2081} to predict the mortality rate of diabetic ICU patients through severity indicators. The application case of this pattern recognition was completely written by Julia language. Other typical applications of pattern recognition techniques are automatic speech recognition, text classification , face recognition.  \textbf{Languages.jl} \cite{2082} is a Julia package for working with human languages. Script detection model works by checking the Unicode character ranges present within the input text. But the package was supported only for English and German currently.

\section{Julia in Machine Learning: Open Issues}

\subsection{Overview}

\begin{figure}[!ht]
	\centering
	\includegraphics[width=0.8\textwidth]{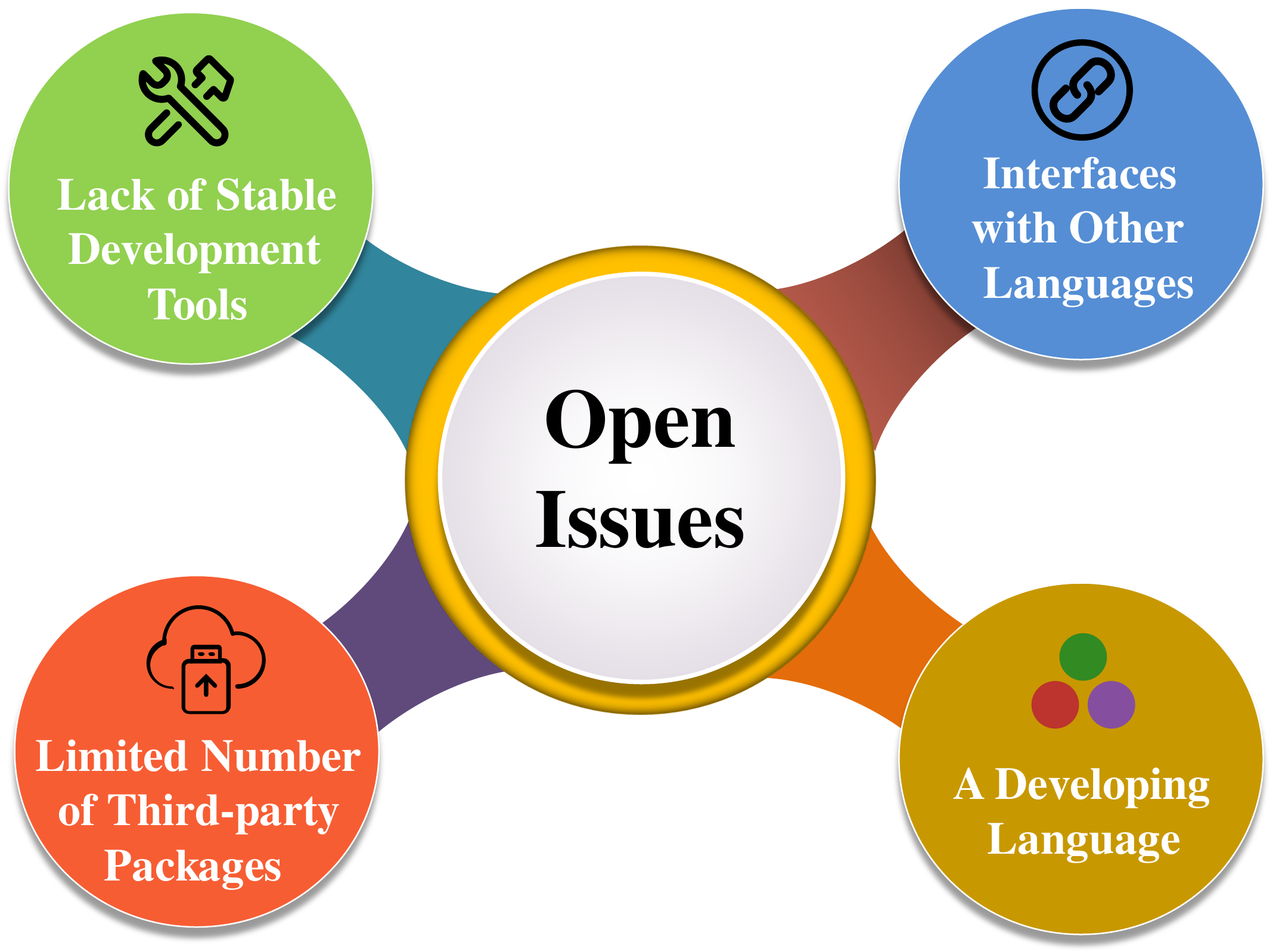}
	\caption{Open issues of Julia language} \label{fig7}
\end{figure}

Since its release, the advantages of Julia language, such as simplicity and efficiency, have been recognized by developers in various fields. However, with the promotion of the Julia language and the steady increase in the number of users, it also faces several open issues; see Figure \ref{fig7}.

\subsection{A Developing Language}

Although Julia has developed rapidly, its influence is far less than that of other popular programming languages. After several versions of updates, Julia has become relatively stable, but there are still several problems to be solved. Julia's grammar has changed considerably, and although these changes are for the sake of performance or ease of expression, these differences also make it difficult for different programs to work together. One of Julia's obvious advantages is its satisfactory efficiency; but to write efficient code, one needs to transform the method of thinking in programming and not just copy code into Julia. For people who have just come into contact with Julia, the ease of use can also cause them to ignore this problem and ultimately lead to unsatisfactory code efficiency.

\subsection{Lack of Stable Development Tools}

Currently, the commonly used editors and IDEs for the Julia language include (1) Juno (Atom Plugin), (2) Visual Studio Code (VS Code Extension), (3) Jupyter (Jupyter kernel), and (4) Jet Brains (IntelliJ IDEA Plugin). According to \cite{124}, Juno is currently the most popular editor. These editors and IDEs are extensions based on third-party platforms, which can quickly build development environments for Julia in its early stages of development, but in the long run, this is not a wise approach. Users need to configure Julia initially, but the final experience is not satisfactory. Programming languages such as MATLAB, Python and C/C++ each have their own IDE, which integrates the functions of code writing, analysis, compilation and debugging. Although editors and IDEs have achieved many excellent functions, it is very important to have a complete and Julia-specific IDE. 

\subsection{Interfacing with Other Languages}

In the process of using Julia for development, although most of the code can be written in Julia, many high-quality, mature numerical computing libraries have been written in C and Fortran. To facilitate the use of existing code, Julia should also make it easy and effective to call C/C++ and Fortran functions. In the field of machine learning, Python has been used to write a large quantity of excellent code. If one desires to transplant code into Julia in a short time for maintenance, simple calls are necessary, which can greatly reduce the learning curve.

Currently, \textit{PyCall} and \textit{Call} are used in Julia to invoke these existing libraries, but Julia still needs a more concise and general invocation method. More important is finding a method to ensure that the parts of these calls can maintain the original execution efficiency or the execution efficiency of the native Julia language. At the same time, it is also very important to embed Julia's code in other languages, which would not only popularize the use of Julia more quickly but also combine the characteristics of Julia to enable researchers to accomplish tasks more quickly.

\subsection{Limited Number of Third-party Packages}

For good programming languages, the quantity and quality of third-party libraries are very important. For Python, there are 194,934 projects registered in \textit{PyPI} \cite{11}, while the number of Julia third-party packages registered in Julia Observer is approximately 3,000. The number of third-party libraries in Julia is increasing, but there are still relatively few compared with other programming languages, and there may not be suitable libraries available in some unpopular areas.

Because Julia is still in the early stage of development, version updates are faster, and the interface and grammar of the program are greatly changed in each version upgrade. After the release of Julia 1.0, the Julia language has become more mature and stable than in the past. However, many excellent third-party machine learning libraries were written before the release of Julia 1.0 and failed to update to the new version in time. Users need to carefully evaluate whether a third-party library has been updated to the latest version of Julia to ensure its normal use. In addition, Julia is designed for parallel programming, but there are not many third-party libraries for parallel programming. Currently, the more commonly used third-party packages of Julia are \textbf{CUDAnative.jl}, \textbf{CuArrays.jl}, and \textbf{juliaDB.jl}. However, many functions in these packages are still in the testing stage.

Although Julia libraries are not as rich as those of Python, the prospects for development are optimistic. Officials have provided statistical trends in the number of repositories. Many scholars and technicians are committed to improving the Julia libraries. Rong Hongbo et al. \cite{125} used Julia, Intel MKL and the SPMP library to implement \textbf{Sparso}, which is a sparse linear algebra context-driven optimization tool that can accelerate machine learning algorithms. Plumb Gregory et al. \cite{126} compiled a library package for fast Fourier analysis using Julia, which made it easier for fast Fourier analysis to be employed in statistical machine learning algorithms.

\section{Conclusions}

This paper has systematically investigated the development status of the Julia language in the field of machine learning, including machine learning algorithms written in Julia, the application of the Julia language in machine learning, and the potential challenges faced by Julia. We find that: (1) Machine learning algorithms written in Julia are mainly supervised learning algorithms, and there are fewer algorithms for unsupervised learning. (2) The Julia language is widely used in seven popular machine learning research topics: pattern recognition, NLP, IoT data analysis, computer vision, autonomous driving, graph analytics, and signal processing. (3) There are far fewer available application packages than there are for other high-level languages, such as Python, which is Julia's greatest challenge. This survey provides a comprehensive investigation of the applications of Julia in machine learning. We believe that with the gradual maturing of the Julia language and the development of related third-party packages, the Julia language will be a highly competitive programming language for machine learning.

\section*{Declaration of competing interest}
The authors declare that they have no known competing interests.

\section*{Acknowledgments}
This research was jointly supported by the National Natural Science 
Foundation of China (11602235), the Fundamental Research Funds for China Central Universities (2652018091), and the Major Project for Science and Technology (2020AA002). The authors would like to thank the editor and the reviewers for their valuable comments.

\newpage
\section*{Reference}
  \bibliographystyle{elsarticle-num} 
  \bibliography{reference}
  

\begin{landscape}
	
\begin{table*}[!h]
	\renewcommand\arraystretch{1}
	\scriptsize  
	\caption{Commonly used Julia language packages}
	\centering
	\begin{tabular}{|c|c|c|c|c|c|c|c|c|c|c|c|c|c|c|c|}
		\hline
		\multirow{2}{*}{\textbf{Julia Packages}} & \multirow{2}{*}{\textbf{Ref.}} & \multicolumn{5}{c|}{\textbf{Supervised Learning}}   & \multicolumn{6}{c|}{\textbf{Unsupervised Learning}} & \multicolumn{3}{c|}{\textbf{Other}} \\ \cline{3-16} 
		&                                & \begin{tabular}[c]{@{}c@{}}Bayesian  \\ Model\end{tabular} & \textit{k}NN & \begin{tabular}[c]{@{}c@{}}Random  \\ Forest\end{tabular} & SVM & \begin{tabular}[c]{@{}c@{}}Regression   \\ Analysis\end{tabular} & GMM & \textit{k}-means & \begin{tabular}[c]{@{}c@{}}Hierarchical   \\ Clustering\end{tabular} & \begin{tabular}[c]{@{}c@{}}Bi- \\ Clustering\end{tabular} & PCA & ICA     & \begin{tabular}[c]{@{}c@{}}Deep \\ Learning \end{tabular} & ANN & ELM  \\ \hline
		ForneyLab.jl                                 &                \cite{24}                & \textbf{\checkmark}    &                &               &                &           & & & &               &            &            &            &                         &               \\ \hline
		NearestNeighbors.jl                             &                 \cite{26}                 & \textbf{}    & \textbf{\checkmark}      & \textbf{}     & \textbf{}      & \textbf{}      & & & &          & \textbf{}  &            &            &                         &               \\ \hline
		DecisionTree.jl                              &             \cite{33}                     & \textbf{}  &   & \textbf{\checkmark}     & \textbf{}     & \textbf{}      & \textbf{}      & & &          & \textbf{}  &            &            &                         &               \\ \hline
		SVM.jl                         &                  \cite{37}                & \textbf{}     & \textbf{}  &    & \textbf{\checkmark}    & \textbf{}      & \textbf{}   & & &             & \textbf{}  &            &            &                         &               \\ \hline
		MLJ.jl                                   &                  \cite{38}                & \textbf{\checkmark}     & \textbf{}      & \textbf{\checkmark}     & \textbf{\checkmark}     & \textbf{\checkmark}      & & & &         & \textbf{}  &            &            &                         &               \\ \hline
		LIBSVM.jl                      &              \cite{39}                    & \textbf{}     & \textbf{}      & \textbf{}     & \textbf{\checkmark}     & \textbf{}   & & & &             & \textbf{}  &            &            &                         &               \\ \hline
		ScikitLearn.jl                                    &                 \cite{40}                 & \textbf{}     & \textbf{}      & \textbf{}     & \textbf{\checkmark}      & \textbf{}               & \textbf{\checkmark}  &       & & & &     &            &                         &               \\ \hline
		Regression.jl                                    &            \cite{46}                      & \textbf{}     & \textbf{}      & \textbf{}     & \textbf{}      & \textbf{\checkmark}      & & & &         & \textbf{} &            &            &                         &               \\ \hline
		EmpiricalRisk.jl                                 &               \cite{47}                   & \textbf{}     & \textbf{}      & \textbf{}     & \textbf{}      & \textbf{\checkmark}   & & & &             & \textbf{} &            &            &                         &               \\ \hline
		Clustering.jl                            &             \cite{55}                     &               &                &               &                &                          & \textbf{} & \textbf{\checkmark} & & & & & \textbf{}  & \textbf{}               & \textbf{}     \\ \hline
		QuickShiftClustering.jl                             &              \cite{61}            &               &                &               &                &                          & \textbf{}  & \textbf{} & \textbf{\checkmark}  & \textbf{}   & & &  &            & \textbf{}     \\ \hline
		kpax3.jl                         &           \cite{8}                       &               &                &               &                &                          & \textbf{}  & \textbf{} & \textbf{}  & \textbf{\checkmark}      & & & &         & \textbf{}     \\ \hline
		MultivariateStats.jl                             &                    \cite{66}              &               &                &               &                &                          & \textbf{}  & \textbf{}  & \textbf{} & \textbf{}      & \textbf{\checkmark}         & \textbf{\checkmark}   & & &  \\ \hline
		MXNet.jl                   &              \cite{79}                    &               &                &               &                &                          & \textbf{}  & \textbf{}  & \textbf{}  & & & & \textbf{\checkmark} &             & \textbf{}     \\ \hline
		Knet.jl                                  &              \cite{80}                    &               &                &               &                &                          & \textbf{}  & \textbf{}  & \textbf{}  & \textbf{}       & &   & \textbf{\checkmark}     & \textbf{\checkmark} &   \\ \hline
		Flux.jl                                    &               \cite{81}                   & \textbf{}     & \textbf{}     & \textbf{}    & \textbf{}      & \textbf{}                & \textbf{} & \textbf{} &            &               &          &     & \textbf{\checkmark} & \textbf{\checkmark}   & \textbf{\checkmark}      \\ \hline
		TensorFlow.jl                                   &            \cite{82}                      & \textbf{}    & \textbf{}      & \textbf{}     & \textbf{}     & \textbf{}                & \textbf{}  & \textbf{}  &            &     &     & &  \textbf{\checkmark}   &                &               \\ \hline
		Diffiqflux.jl                                    &            \cite{86}                      & \textbf{}    & \textbf{}      & \textbf{}     & \textbf{}     & \textbf{}                & \textbf{}  & \textbf{}  &            &    & &      & &  \textbf{\checkmark}   &                            \\ \hline
		DifferentialEquations.jl                                     &            \cite{87}                      & \textbf{}    & \textbf{}      & \textbf{}     & \textbf{}     & \textbf{}                & \textbf{}  & \textbf{}  &            &    & &      & &  \textbf{\checkmark}   &                            \\ \hline
		Backpropneuralnet.jl                                    &            \cite{52}                      & \textbf{}    & \textbf{}      & \textbf{}     & \textbf{}     & \textbf{}                & \textbf{}  & \textbf{}  &            &    & &      & &  \textbf{\checkmark}   &                            \\ \hline
		Elm.jl                                     &            \cite{91}                      & \textbf{}    & \textbf{}      & \textbf{}     & \textbf{}     & \textbf{}                & \textbf{}  & \textbf{}  &            &    & &   &   & &  \textbf{\checkmark}                              \\ \hline
		GmmFlow.jl                                     &            \cite{2051}                      & \textbf{}    & \textbf{}      & \textbf{}     & \textbf{}     & \textbf{}                &\textbf{\checkmark}  & \textbf{}  &            &    & &   &   & &  \textbf{}                              \\ \hline
		GaussianMixtures.jl                                     &            \cite{2052}                      & \textbf{}    & \textbf{}      & \textbf{}     & \textbf{}     & \textbf{}                & \textbf{\checkmark}  & \textbf{}  &            &    & &   &   & &  \textbf{}                              \\ \hline
	\end{tabular}
	\label{tab1}
\end{table*}

\end{landscape}

\end{document}